\documentclass[runningheads]{llncs}
\usepackage{url}

\usepackage{todonotes}
\usepackage{subfigure}

\usepackage[acronym]{glossaries}
\newacronym{ALS}{ALS}{airborne laser scanning}
\newacronym{MLS}{MLS}{mobile laser scanning}
\newacronym{LoD}{LoD}{Level of Detail}
\newacronym{OGC}{OGC}{Open Geospatial Consortium}
\newacronym{GML}{GML}{Geography Markup Language}
\newacronym{ASAM}{ASAM}{Association for Standardization of Automation and Measuring Systems}
\newacronym{TLS}{TLS}{terrestrial laser scanning}
\newacronym{UAV}{UAV}{unmanned aerial vehicle}
\newacronym{HD}{HD}{high definition}
\newacronym{RANSAC}{RANSAC}{RANdom SAmple Consensus}
\newacronym{ROI}{ROI}{region of interest}
\newacronym{DEM}{DEM}{digital elevation model}
\newacronym{ICP}{ICP}{iterative closest point}
\newacronym{NLOS}{NLOS}{non-line-of-sight}
\newacronym{SfM}{SfM}{structure from motion}
\newacronym{FME}{FME}{Feature Manipulation Engine}
\newacronym{OSM}{OSM}{OpenStreetMap} 
\newacronym{RMSE}{RMSE}{root mean square error}
\newacronym{CPT}{CPT}{conditional probability table}
\newacronym{DST}{DST}{Dempster–Shafer theory}
\newacronym{BN}{BayNet}{Bayesian network}
\newacronym{GIS}{GIS}{Geographic Information System}
\newacronym{PPD}{PPD}{posterior probability distribution}
\newacronym{CI}{CI}{confidence interval}
\newacronym{LiDAR}{LiDAR}{light detection and ranging}
\newacronym{SVM}{SVM}{support vector machines}
\newacronym{ML}{ML}{machine learning}
\newacronym{DL}{DL}{deep learning}
\newacronym{PCA}{PCA}{principal component analysis}
\newacronym{KD}{KD}{K-Dimension}
\newacronym{SVD}{SVD}{Singular Value Decomposition}
\newacronym{KNN}{KNN}{k-Nearest Neighbor}
\newacronym{MLP}{MLP}{multilayer perceptron}
\newacronym{DGCNN}{DGCNN}{Dynamic Graph CNN}

\usepackage{graphicx}

\begin{document}
\title{Classifying point clouds at the facade-level using geometric features and deep learning networks}
\titlerunning{Classifying point clouds at the facade-level using GF and DL}
\author{Yue Tan\textsuperscript{1 }, Olaf Wysocki\textsuperscript{1 }, Ludwig Hoegner\textsuperscript{1,2 }, Uwe Stilla\textsuperscript{1 }}
\authorrunning{Tan Y., et al.,}
%
 \institute{Photogrammetry and Remote Sensing, TUM School of Engineering and Design, Technical University of Munich (TUM),\\ Munich, Germany - (yue.tan, olaf.wysocki, ludwig.hoegner, stilla)@tum.de \and Department of Geoinformatics, University of Applied Science (HM), Munich, Germany - ludwig.hoegner@hm.edu\\}
%
\maketitle              

\begin{abstract}
3D building models with facade details are playing an important role in many applications now.
Classifying point clouds at facade-level is key to create such digital replicas of the real world.
However, few studies have focused on such detailed classification with deep neural networks.
We propose a method fusing geometric features with deep learning networks for point cloud classification at facade-level. 
Our experiments conclude that such early-fused features improve deep learning methods' performance.
This method can be applied for compensating deep learning networks' ability in capturing local geometric information and promoting the advancement of semantic segmentation.

\end{abstract}
\section{Introduction}
Nowadays, semantic 3D building models are widely used in many fields, such as architecture, engineering, construction, and facilities management~\cite{biljecki2015applications}.
Meanwhile, the widespread use of~\gls{LiDAR} data provides the possibility of at-scale reconstruction of 3D building models up to~\gls{LoD}2~\cite{haala2010update}~\footnote{https://github.com/OloOcki/awesome-citygml}.
Street-level point clouds with rich and detailed facade-level semantic can enable the reconstruction of highly detailed~\gls{LoD}3 building models.
In addition, the reliable and detailed segmentation of point clouds can also provide improvement on the achievable~\gls{LoD}. 
Such point clouds stem from~\gls{MLS} units, whose availability has increased in recent years~\cite{Wysocki2022}.
With the wide use of low-cost laser scanning system, it is easy and cheaper to acquire point clouds now. 
However, the automatic interpretation of 3D point clouds by semantic segmentation for high~\gls{LoD} reconstruction represents a very challenging task~\cite{Grilli2020}. 
It leads to a growing need of innovative point cloud classification methods to extract highly-detailed semantic information~\cite{Grilli2020}.

In this study, we propose a method combining the deep learning networks and geometric features together for improving networks’ performance in point cloud classification at facade-level. 
Our contributions are as follows:
\begin{itemize}
    \item [$\bullet$] Comparison of~\gls{DL} and~\gls{ML} approaches on classifying building point clouds at facade-level
    \item [$\bullet$] A method improving the performance of~\gls{DL} networks by adding geometric features into~\gls{DL} approaches
    \item [$\bullet$] Analysis on impact from selection of geometric features upon the deep learning networks
\end{itemize}

\section{Related Works}
Recently, machine- and its subset deep-learning algorithms (\gls{ML} / \gls{DL}) have become the state-of-the-art approach to classify point clouds~\cite{Grilli2020}. 
For example. the traditional machine-learning-based approaches that apply geometric features for point cloud classification, such as Random Forest, have been used in the cultural heritage domain~\cite{Grilli2020}. 
Another example comes from Weimann et al., who employ urban scenario point cloud with traditional \gls{ML} models including \gls{KNN} and \gls{SVM} using geometric features \cite{Weinmann2013}.
In addition to these mentioned~\gls{ML} algorithms, 3D neural networks also have exhibited formidable capabilities in point cloud segmentation.
PointNet \cite{Qi_2017_CVPR} and its successor PointNet++ \cite{qi2017pointnet++} have achieved promising results in the field of point cloud classification, especially for man-made objects.  
Besides, the utilization of the self-attention mechanism in point cloud classification also has yielded remarkable classification outcomes, and the Point Transformer algorithm emerges as an illustrious paradigm of its application~\cite{Zhao_2021_ICCV}.
Moreover, the graph-based method \gls{DGCNN} has been proven to attain the new state of the art performance in roadway target detection for point cloud dataset~\cite{Simonovsky_2017_CVPR}.

In general, the difference between \gls{DL} approaches and other machine learning approaches is that features are learned from the training process.
This kind of representation learning can provide access to capture undefined structure information of point cloud in training, which is often seen as the reason for the rapid development in 2D and 3D understanding tasks~\cite{griffiths2019review}.

Increasing number of deep learning methods are proposed to process point cloud data, however, only few studies are devoted to point cloud segmentation at facade-level for~\gls{LoD}3 reconstruction. 
The complexity of the architecture scene makes it a challenging task to classify buildings’ point clouds with rich semantics information, especially for distinguishing facade components that are translucent (e.g., windows) or with low ratio to the overall point cloud (e.g., doors)~\cite{Wysocki2022}. 
\section{Methodology}
\begin{figure}
\centering
\includegraphics[width=0.9\textwidth]{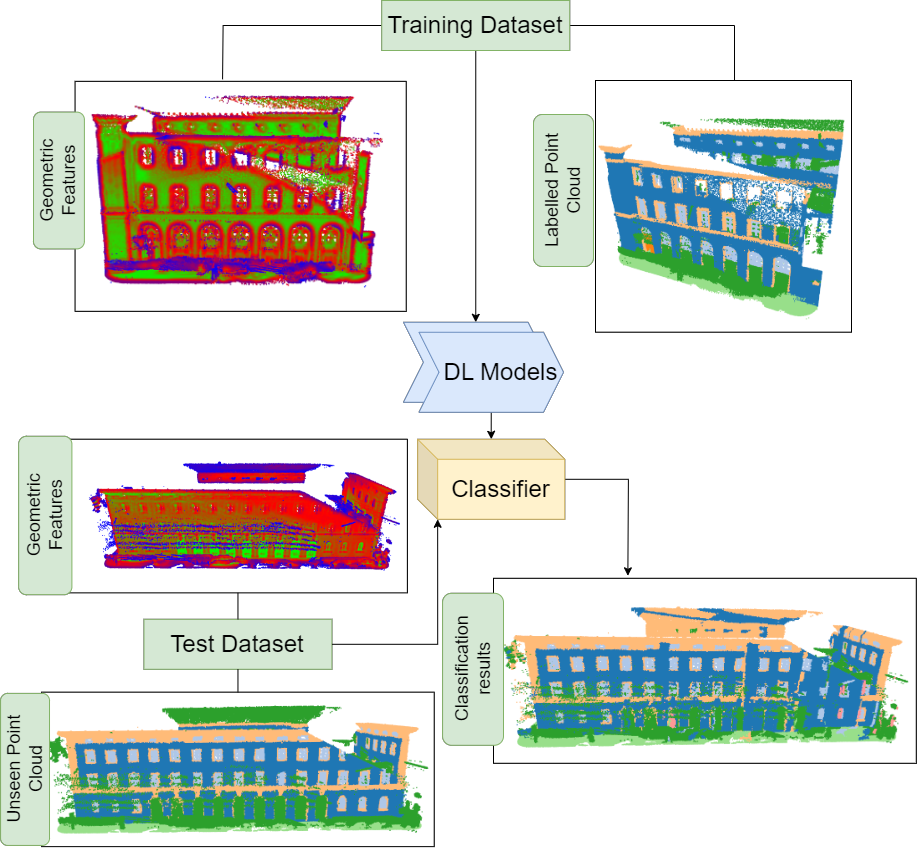}
\caption{Point cloud classification with geometric features process for unseen datasets}
\label{Classification_Process}
\end{figure}

In this paper, we focus on improving the neural networks' classification performance for point cloud using geometric features.
For this purpose, as shown in Figure~\ref{Classification_Process}, both point coordinates and geometric features are prepared for training the selected~\gls{DL}
models.
With the classifier, performance on unseen scenarios can be validated.

\subsection{Geometric features extraction}
The careful selection for compact and robust geometric features subset is based on previous studies.
There have been many studies demonstrating the powerful capacity of geometric features in point cloud classification, which can be viewed as a detailed semantic interpretation of local 3D structures~\cite{Grilli2020,Weinmann2013}. 
According to the study of Grill and Remondino~\cite{Grilli2020}, $planarity$(p), $omnivariance$(o), and $surface~variation$(c) are efficient in classification of building point cloud data applying Random Forest.
In this paper, except for the mentioned geometric features, we also include \gls{PCA} variables and its eigenvector for each point in classifying the points. 
The selected covariance geometric features are used for describing local structure's dimension and shape.
Here, $planarity$ can provide information about the presence of planar 3D structure, while $surface~variation$ measures the change of curvature for the local structure \cite{Weinmann2013}.
Besides, \gls{PCA} components and its further measurement $omnivariance$ obtained by the covariance matrix can summarize the distribution of surrounding points within the local window, and eigenvectors are able to show the direction of such a cluster's distribution.
These geometric features exhibit strong responsiveness to the buildings' detailed facade in Grilli's study~\cite{Grilli2020}, such as protruding moulding and open windows, showing their potential in enhancement for \gls{DL} models.

\begin{figure}
\centering
\includegraphics[width=0.9\textwidth]{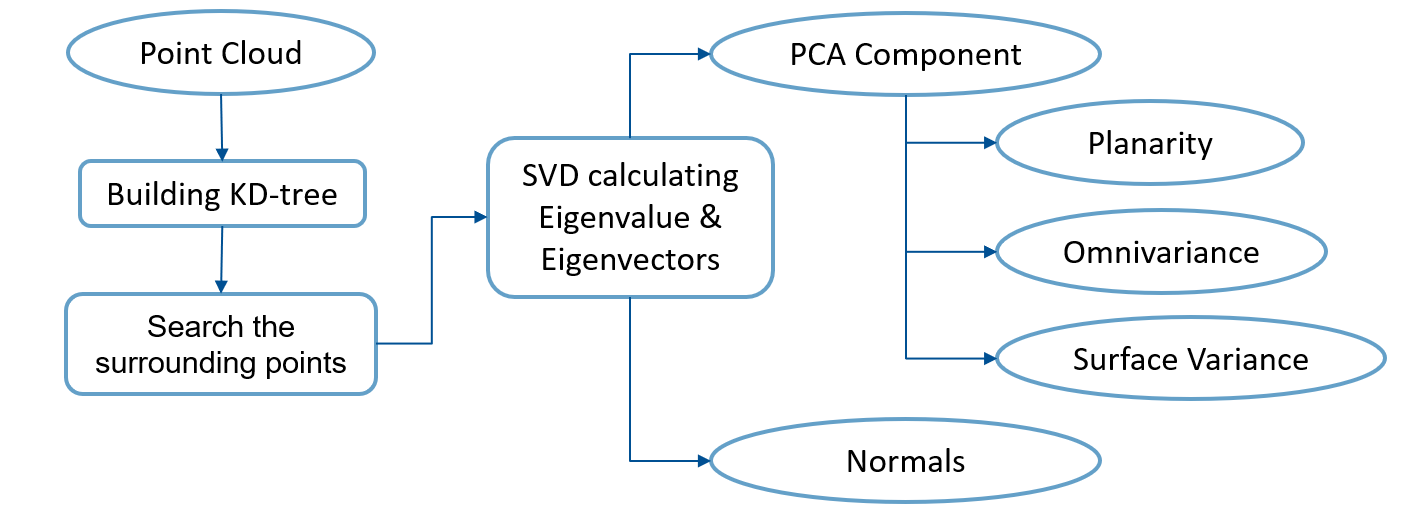}
\caption{Workflow of geometric features extraction}
\label{GF_Pipeline}
\end{figure}

In our method, geometric features are calculated based on surrounding points that locate in a spherical space with a fixed radius. 
To find such surrounding points for each point, we apply a \gls{KD} tree for nearest neighbor queries. 
Then, we calculate the respective structure tensor based on \gls{SVD}, which can directly provide information of surrounding points' distribution and structure.
In this way, the eigenvalues $\lambda_1$, $\lambda_2$, $\lambda_3$, as well as the corresponding eigenvectors $e_1$, $e_2$, $e_3$ are derived from the covariance matrix. 
In addition, the eigenvalues with $\lambda_1 > \lambda_2 > \lambda_3$, known as \gls{PCA} components, are used for further measures of other covariance features~\cite{Weinmann2013}:
\begin{eqnarray}
    p = \frac{\lambda_2-\lambda _1}{\lambda_1},
\end{eqnarray}
\begin{eqnarray}
    o = (\lambda_1 \times \lambda_2 \times \lambda_3)^{1\over3}, 
\end{eqnarray}
\begin{eqnarray}
    c = \frac{\lambda_3}{\lambda_1+\lambda_2+\lambda_3}
\end{eqnarray}
Besides, the second vector of eigenvectors, which can describe the general direction perpendicular to the curve defined by surrounding points, is also included.
Figure \ref{GF_Pipeline} represents how the geometric features are calculated.
\subsection{Classification}
Our method employs two deep learning models PointNet and PointNet++, in which the geometric features are early-fused. 
In addition, result of Random Forest is also evaluated as reference not only for the deep neural networks but also for the selection of geometric features.
Point-based networks represent the most extensively examined approaches within the realm of academic research, among which PointNet is considered as pioneer work.
It learns pointwise features separately for each point using \gls{MLP} layers and aggregates a global representation for the overall batch of  point cloud with a max-pooling layer~\cite{Guo_review}.
While PointNet++ is an improved version of PointNet, it improves the limitation that PointNet cannot capture local structures in different metric space where points exist~\cite{qi2017pointnet++}. 
With a hierarchical structure designed inside the networks, PointNet++ has the capability to extract information regarding local geometric features.

\begin{figure}
\centering
\includegraphics[width=0.9\textwidth]{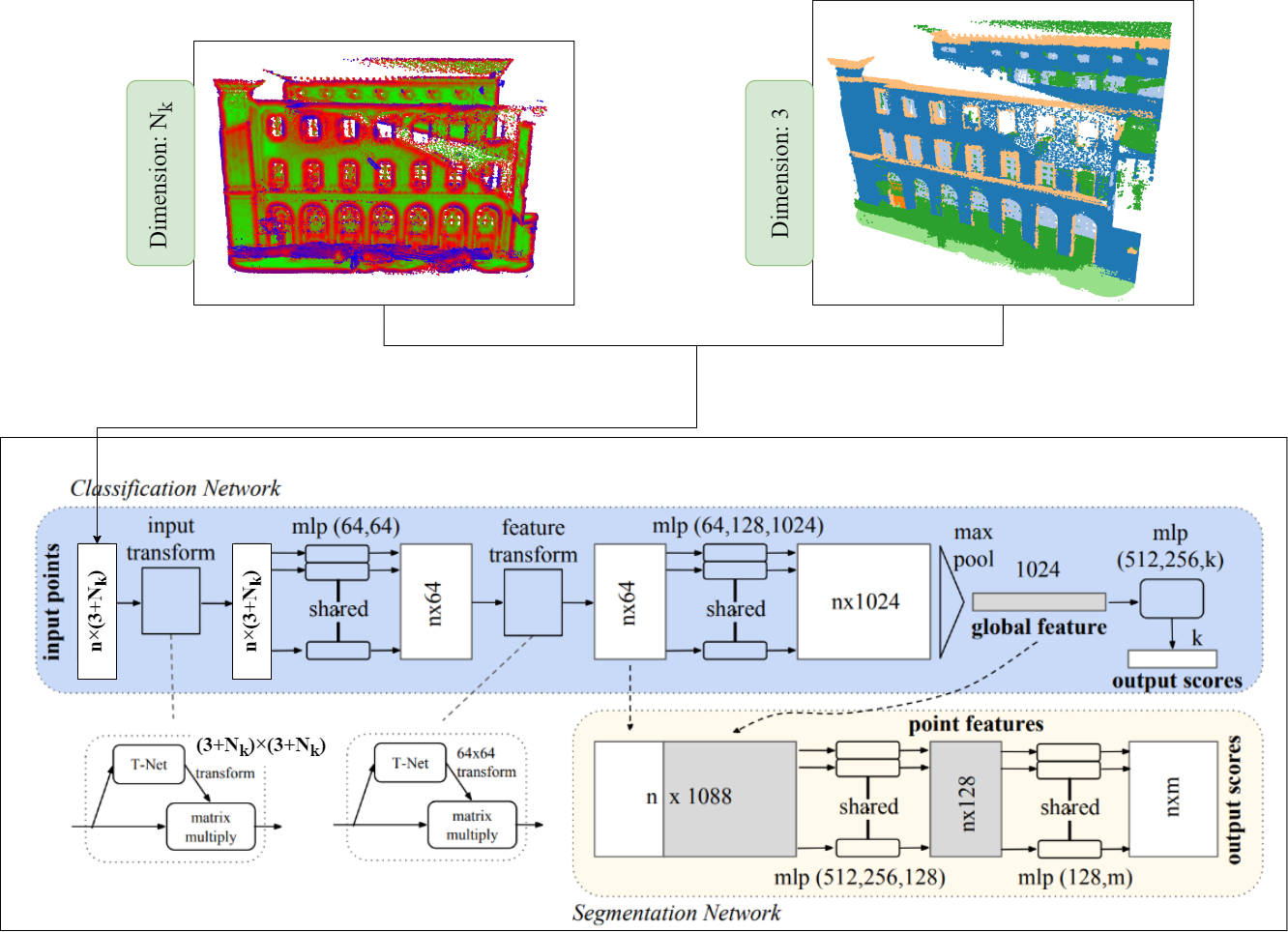}
\caption{Proposed early-fusion in PointNet networks. Created based on~\cite{Qi_2017_CVPR} }
\label{GF_Networks}
\end{figure}

In order to add geometric features into networks, internal structure of the neural networks needs to be updated.  
For PointNet, features learned in this network are calculated by convolution on coordinate of each point and then gathered globally into a general single layer.
Hence, we expand the input dimension of coordinate into $3+N_f$, where $N_f$ is the number of pre-computed geometric features.
As for PointNet++, features from different scale of metric spaces are collected using hierarchical structure so-called set abstraction layers, consisting of sample layer, grouping layer and PointNet layer. 
The former two layers generate regions in different scales using point cloud coordinates, while the PointNet layer is used for gathering features from group of points in different regions created in the former two layers. 
For this reason, we increase the input dimension for PointNet layer to $N \times K \times (d+N_f) $, where N is the number of points, K is the number of regions after grouping, d is the dimension of coordinates, $N_f$ is the dimension of external geometric features. We demonstrate such an early-fusion in Figure~\ref{GF_Networks}. 

Apart from experiment with these~\gls{DL} models, performance of the~\gls{ML} model Random Forest is also tested, and importance of all dimensions from geometric features as well as coordinates is determined. 
With this result as a prior knowledge, different combinations of geometric features are selected and applied for improving the deep learning networks’ performance.

\section{Experiments}
\subsection{Datasets}
\begin{figure}
\centering
\includegraphics[width=0.9\textwidth]{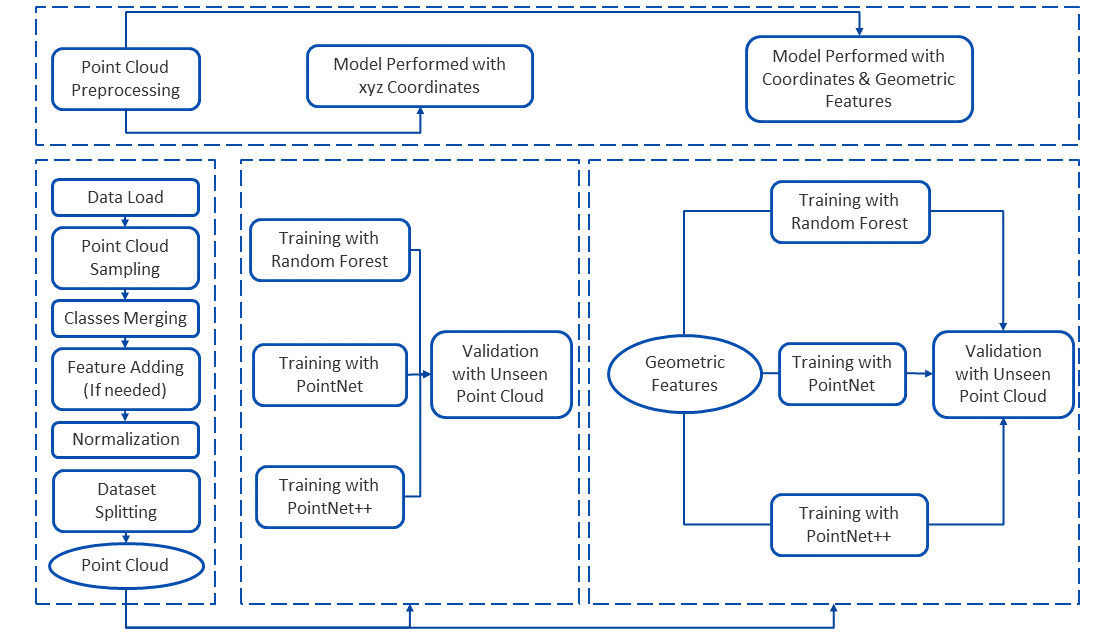}
\caption{Experiments set-up}
\label{Pipeline}
\end{figure}

The TUM-FAÇADE dataset was used in this experiment comprising~\gls{MLS} point clouds representing the main campus of the Technical University of Munich, Munich, Germany~\cite{wysocki2023tumfacadeArxiv}. 
We tested the performance on two buildings: No.4959323 and No.4959459, and different training datasets were separately applied for the two unseen buildings. 
Here, unseen building refers to dataset that would not participate in training the~\gls{DL} models but shall be used for testing the performance of the classifiers.
For the point cloud of building No.4959323, four buildings including No.4906981, No.4959322, No.4959460 and No.4959462 were gathered as the training data. 
While for the building No.4959459, the building No.4959322 was set as the training data. 
For pre-processing, we downsampled and merged 17 classes into 7 representative facade classes considering the complexity of the classification, unbalanced contribution of points in classes, following the approach of Wysocki et al.,~\cite{Wysocki2022}. 
The original manually labelled classes include $wall$, $window$, $door$, $balcony$, $molding$, $deco$(decoration), $column$, $arch$, $stair$, $ground~surface$, $terrain$, $roof$, $blinds$, $outer~ceiling~surface$, $interior$ and $other$, after pre-processing they were merged into $wall$, $molding$, $door$, $column$, $arch$, $terrian$, $roof$ and $other$. 
For validating the building No.4959323, redundant points in training data were decreased significantly after downsampling within a distance of 0.1m, from 81 million to 4 million, while for building No.4959459 the sampling distance was set as 0.05m, number of points was reduced from 5 million to 1 million.
Experiment set-up is shown in Figure~\ref{Pipeline}.
\subsection{Geometric features}
\begin{figure}
\centering
\includegraphics[width=0.9\textwidth]{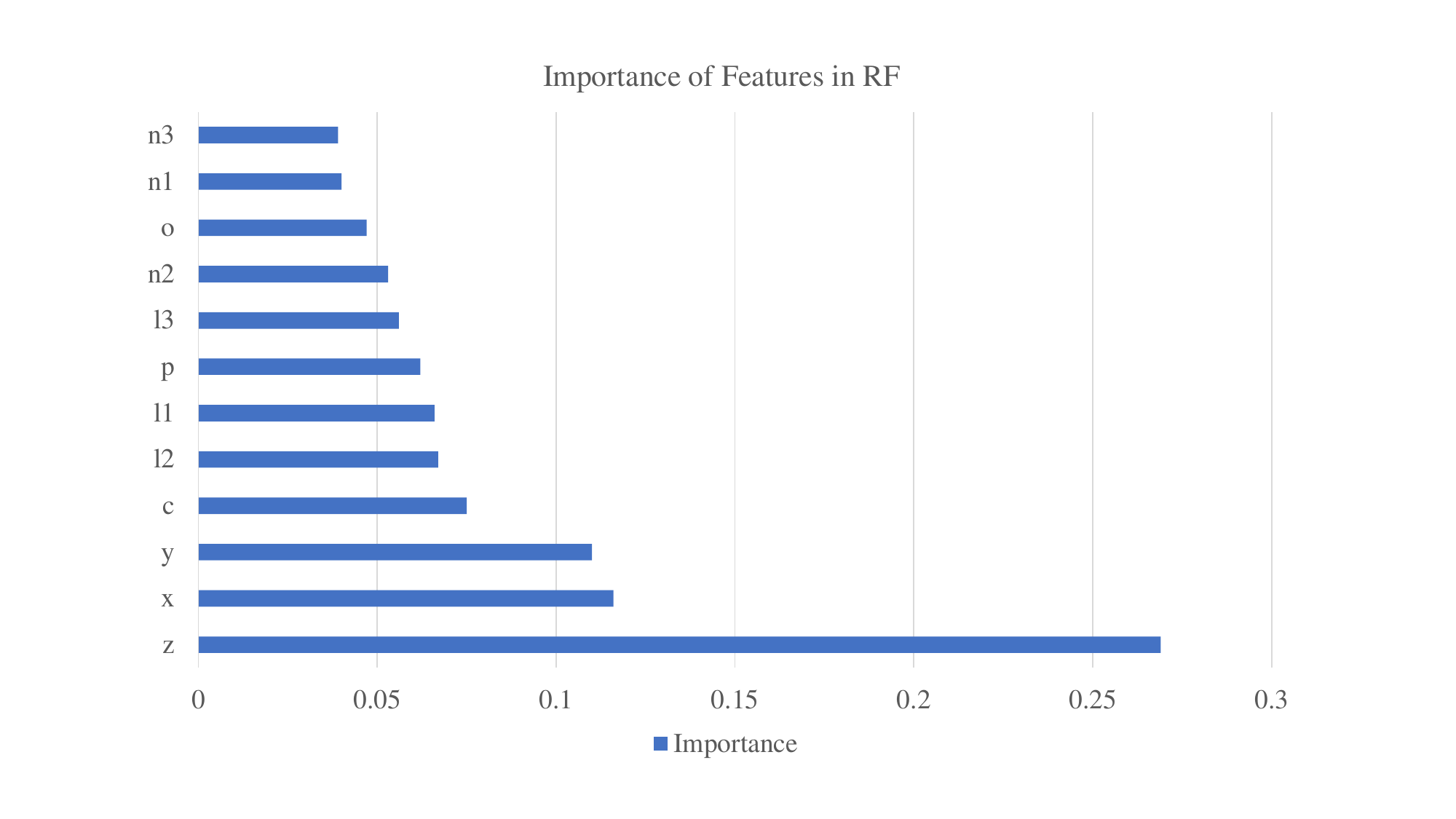}
\caption{Feature importance ranking in Random Forest}
\label{RF_FI}
\end{figure}

\begin{figure}
\centering
\includegraphics[width=0.9\textwidth]{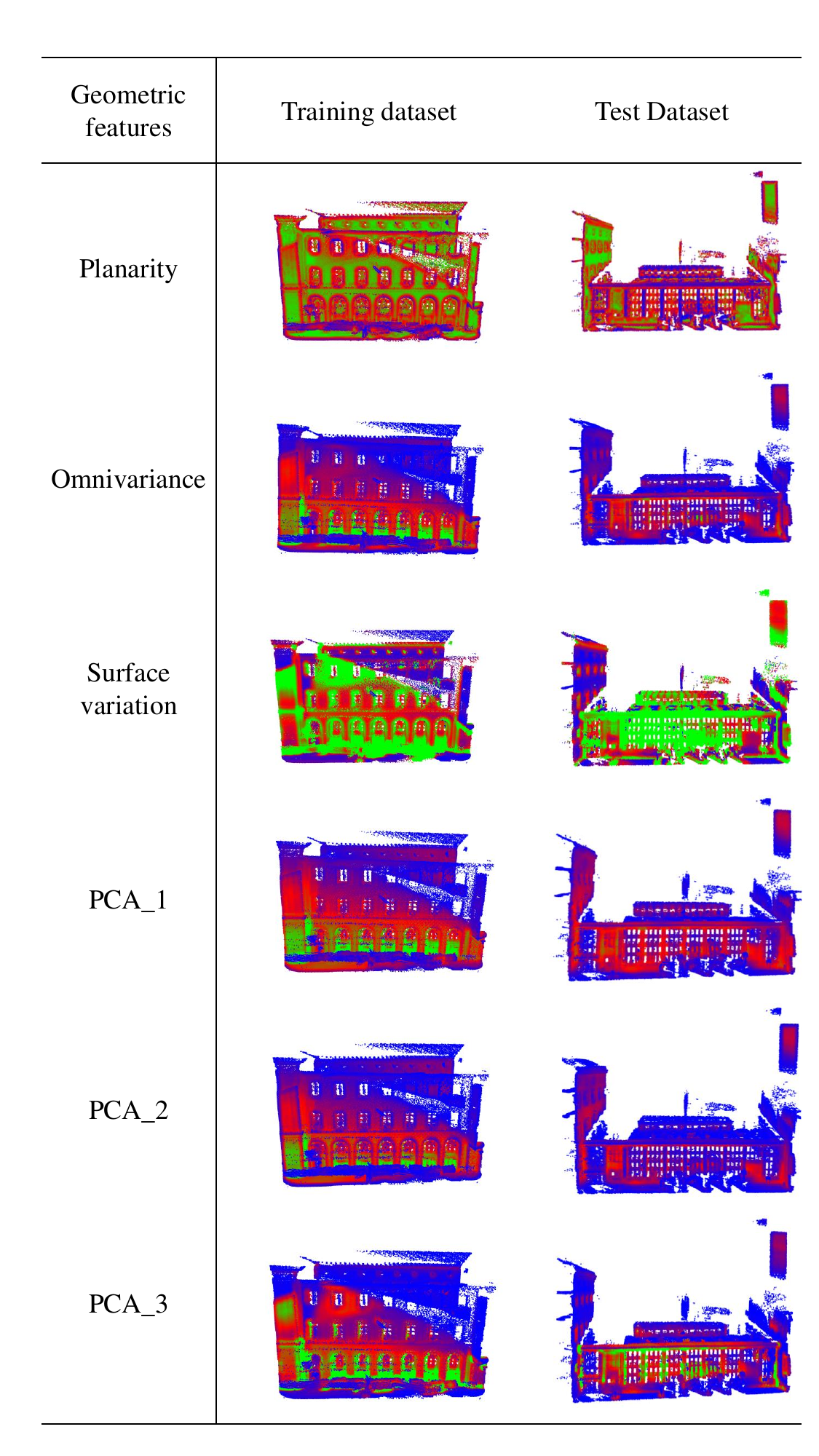}
\caption{Covariance geometric features used to train the \gls{ML} and support \gls{DL} classifiers.}
\label{GF_List}
\end{figure}

Using the method presented in Figure~\ref{GF_Pipeline}, we extracted six covariance geometric features (see Figure~\ref{GF_List}) as well as the desired eigenvector.
To obtain local geometric features, the radius of spherical space for each point was set as 0.8m.
In addition, Random Forest also measured the importance of each feature component, and the result was used for selection of different geometric features combinations.
For analysis on different combination of geometric features, we calculated the importance of different features with Random Forest (Figure~\ref{RF_FI}).
As shown in Figure~\ref{RF_FI}, coordinate components $x$, $y$, $z$ were the most powerful factors in Random Forest classification.
Features with a score over 0.05 were also considered to be influential in this part of experiment, including $surface~variation$, $palnarity$, \gls{PCA} components and the second dimension of second eigenvector.
Based on this result, we selected two different feature combinations, one selection included nine kinds of features as input: $planarity$, $surface~variation$, $omnivariance$, three \gls{PCA} components, and three dimensions of second eigenvector, while for the other, $omnivariance$, the first dimension and third dimension of second eigenvector were removed due to their insignificant performance in Random Forest classification, only top six features in terms of importance including $surface~variation$, $planarity$, \gls{PCA} components and the second dimension of second eigenvector were kept for comparison.
These two selections of geometric features served as an additional input for \gls{DL} models PointNet and PointNet++.

\subsection{Validation of improved models} \label{Re}
\begin{figure}
\centering
\includegraphics[width=0.95\textwidth]{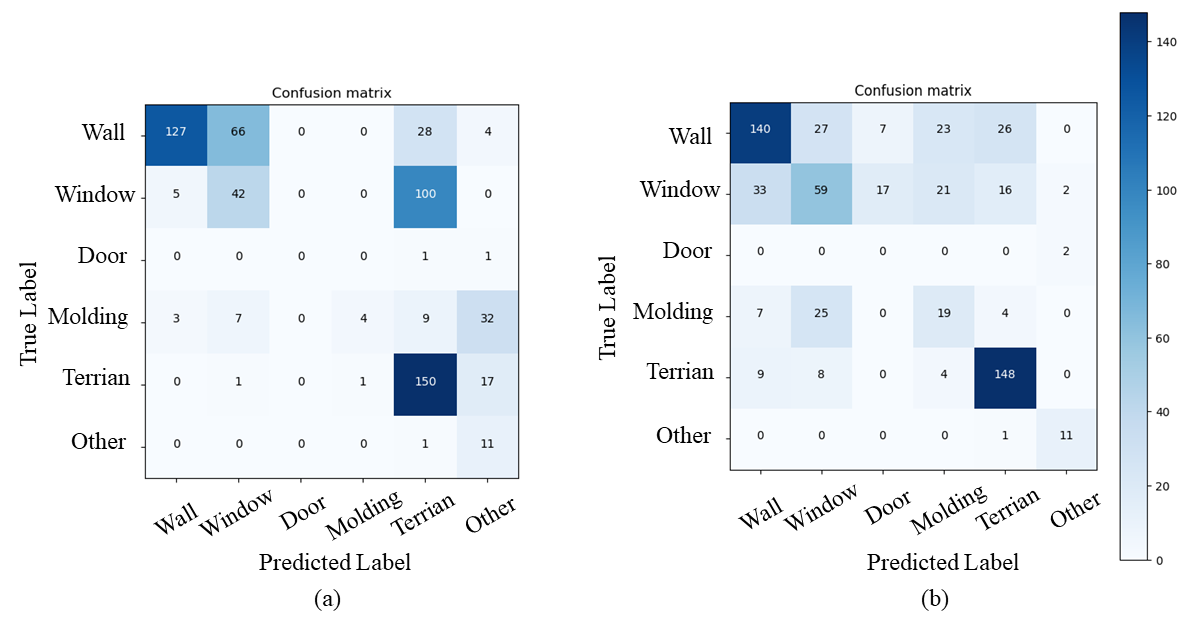}
\caption{PointNet++ classification result on building No.59 with and without geometric features, (a) PointNet++ with xyz only(54.8\%), (b) PointNet++ with xyz and 6 geometric features(62.1\%)}
\label{PointNet2_re_quantity}
\end{figure}

\begin{figure}
\centering
\includegraphics[width=0.95\textwidth]{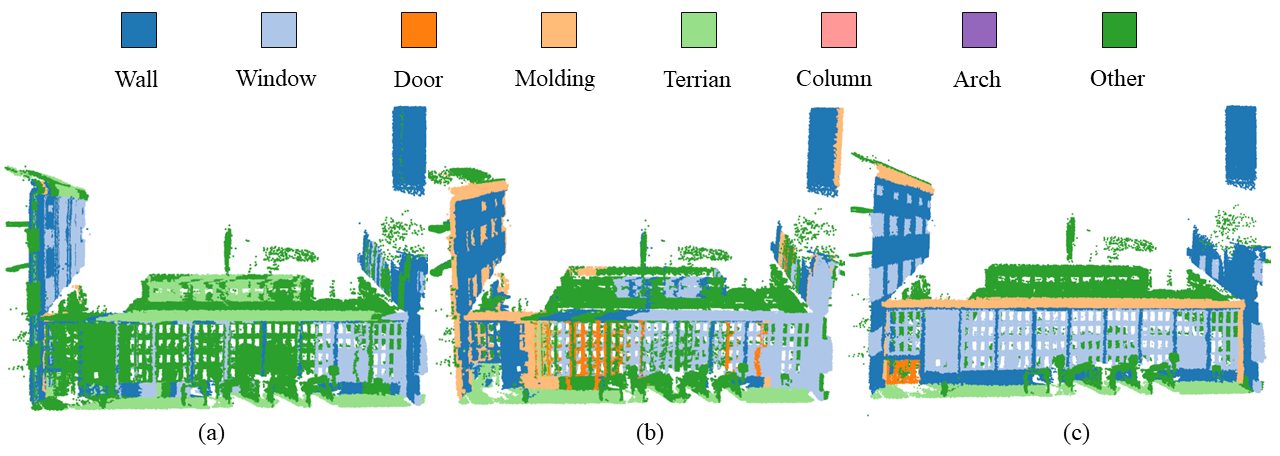}
\caption{PointNet++ classification result on building 59 with and without geometric features, (a) PointNet++ with xyz only(54.8\%), (b) PointNet++ with xyz and 6 geometric features(62.1\%), (c) manually labelled result}
\label{PointNet2_re_quality}
\end{figure}

\begin{figure}
\centering
\includegraphics[width=0.95\textwidth]{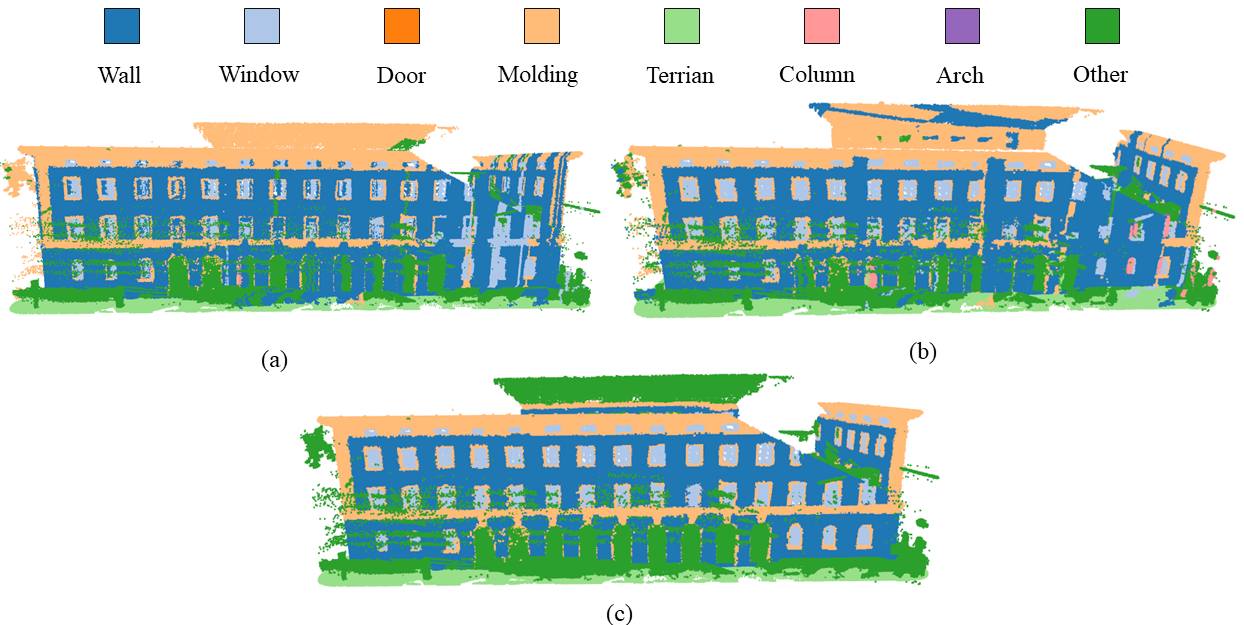}
\caption{PointNet++ classification result on building 23 with and without geometric features, (a) PointNet++ with xyz only(83.1\%), (b) PointNet++ with xyz and 6 geometric features(87.5\%), (c) manually labelled result}
\label{PointNet2_re_quality_23}
\end{figure}

Validation of improved models for facade-level classification was based on the overall accuracy of different approaches. 
Except for comparison between with and without geometric features, different combinations of geometric features were also tested in this study. 
As shown in Table~\ref{Accuracy_OA}, the dataset labelled with 59 refers to performance on the building No.4959459, and 23 means validation on building No.4959323.
In addition to dataset with only coordinates as input (XYZ), dataset with nine kinds of geometric features (XYZ+9F) and six kinds of geometric features (XYZ+6F) were also evaluated.
From statistical results in Table~\ref{Accuracy_OA}, we can see that among all the various combinations, coordinates with six kinds of geometric features (XYZ+6F) as input performs better for PointNet++.
Therefore, we took the result of XYZ and XYZ+6f for visualization analysis on influence of geometric features.
Confusion matrix in Figure~\ref{PointNet2_re_quantity} represents comparison of classification results on different classes using PointNet++ with and without geometric features.
Moreover, a visual examination (Figure \ref{PointNet2_re_quality}, Figure \ref{PointNet2_re_quality_23}) also shows us how PointNet++ performs with or without geometric features on the two unseen buildings.
The implementation is available in the repository~\footnote{https://github.com/yue-t99/PointNet2-GeometricFeatures-Facade}.

\begin{table}
\caption{Overall accuracy for different methods}
\begin{center}
\begin{tabular}{cccc}
\hline
\multicolumn{1}{l}{\rule{0pt}{12pt}Datasets} 
& \multicolumn{1}{l}{RF} & \multicolumn{1}{l}{PointNet} & \multicolumn{1}{l}{PointNet++}\\
\hline\rule{0pt}{12pt}
   59\_XYZ& 33.2\%          & 39.6\%          & 54.8\% \\              
59\_XYZ+9F& 49.7\%          & 55.8\%          & 39.2\% \\
59\_XYZ+6F& 49.1\%          & 52.1\%          & \textbf{62.1\%} \\
   23\_XYZ& 68.4\%          & 69.1\%          & 81.3\% \\              
23\_XYZ+9F& 66.5\%          & 78.5\%          & 84.7\% \\
23\_XYZ+6F& 64.2\%          & 85.5\%          & \textbf{87.5\%} \\[2pt]
\hline
\end{tabular}
\end{center}
\label{Accuracy_OA}
\end{table}
\section{Discussion}
The accuracy results of different models corroborate that adding geometric features do improve the performance of deep neural network. 
As presented in Table~\ref{Accuracy_OA}, overall accuracy with PointNet tends to be divergent when there is only coordinates as input data and one building set as training dataset, and adding geometric features can help to avoid this problem, since more local structure information is introduced to this~\gls{DL} model.
Classification results in PointNet++ represent the same enhancement with six geometric features, respectively from 54.8\% to 62.1\% for the building No.59 and 83.1\% to 87.5\% for building No.23.
Observation from visualized classification results also verifies this statement.
Comparing to PointNet++ trained with coordinates only, the features-extended version in Figure~\ref{PointNet2_re_quality} and Figure~\ref{PointNet2_re_quality_23} presents to be more accurate in distinguishing molding and window.

From the accuracy result in Table~\ref{Accuracy_OA}, we also observe that when applying the same dataset for training, among all the approaches tested, PointNet++ with both coordinates and geometric features gives the most accurate result, with an accuracy of 87.5\%.
Comparing different selection of geometric features, for the deep learning model PointNet++, $surface~variation$, $planarity$, \gls{PCA} components and the second dimension of second eigenvector can improve the performance. 

From the accuracy result tested with different combination of geometric features, we can also reasonably infer that selection of geometric features is important, otherwise accuracy might be decreased. 
In Table~\ref{Accuracy_OA}, when applying combination of nine geometric features for testing on building No.59, instead of improvement in overall accuracy, the performance of PointNet++ is decreased.

In our method, we choose PointNet and PointNet++ due to their different sensitivity in capturing local structures, and improvement for PointNet is more significant than PointNet++. 
In previous studies, PointNet shows limits in recognizing local structure.
By comparing the results of these two~\gls{DL} models, we can have a deeper insight of the improvement from geometric features.
In this experiment, when applying xyz coordinates as input data only, this weakness of PointNet remains noticeable comparing to its' modified version PointNet++.

\section{Conclusion}
Our work demonstrates the capability of geometric features in enhancing deep learning networks' accuracy for facade-level point cloud classification.
The method proposed in this paper can compensate for the inadequacy of deep learning networks in capturing local structure information to a certain extent, providing more sufficient solution for automatic interpretation of 3D buildings.

The validation presents that comparing to the traditional~\gls{DL} and~\gls{ML} method, our early-fused solution reaches a increased accuracy by approximately 10\% for the~\gls{DL} model PointNet and 5\% for PointNet++, leading us to the conclusion that our method could be a alternative supplement for the point-based neural networks.
Such enhancement possesses the potential for further expansion in employing deep learning networks for generalization across large 3D building scenario. 

The assessed facades exhibited intricate and diverse measuring conditions, posing a challenge for classifying and testing~\cite{Wysocki2022} and leading to different levels of sensitivity in various features. 
Furthermore, the deep learning method we used in our experiments relies on a point-based approach that requires a computationally intensive neighborhood search mechanism~\cite{Guo_review}, which duplicates the pre-computed geometric features, leading to a decrease in computational efficiency.
Therefore, employing more different kinds of~\gls{DL} models represents an alternative approach for improvement that can be considered.
With more geometric features applied in, optimized combination for different~\gls{DL} models could be utilized for detailed segmentation of point cloud.
In future, our work will focus on generalizing such experiments for high~\gls{LoD} reconstruction.

Our work provides an opportunity for a better understanding of the detailed facade in building point cloud with deep neural networks, which could be used for many scene understanding related application, such as visualization for navigation purposes~\cite{biljecki2015applications}.

\section{Acknowledgement}
This work was supported by the Bavarian State Ministry for Economic Affairs, Regional Development and Energy within the framework of the IuK Bayern project \textit{MoFa3D - Mobile Erfassung von Fassaden mittels 3D Punktwolken}, Grant No.\ IUK643/001.
Moreover, the work was conducted within the framework of the Leonhard Obermeyer Center at the Technical University of Munich (TUM).
%
%
\bibliographystyle{spbasic}
\bibliography{mybiblio.bib}
\end{document}